\documentclass{article}

\usepackage{microtype}
\usepackage{graphicx}
\usepackage{subcaption}
\usepackage{booktabs} 

\usepackage{hyperref}

\usepackage[accepted]{icml2026}

\usepackage{amsmath}
\usepackage{amssymb}
\usepackage{mathtools}
\usepackage{amsthm}

\usepackage[capitalize,noabbrev]{cleveref}

\theoremstyle{plain}

\theoremstyle{definition}

\theoremstyle{remark}

\usepackage[textsize=tiny]{todonotes}

\icmltitlerunning{Autoreformalization}

\begin{document}

\twocolumn[
  \icmltitle{Reformalization of the Jordan Curve Theorem}



  \begin{icmlauthorlist}
    \icmlauthor{Simon Guilloud}{epfl}
    \icmlauthor{Sankalp Gambhir}{epfl}
    \icmlauthor{Samuel Chassot}{epfl}
  \end{icmlauthorlist}

  \icmlaffiliation{epfl}{École Polytechnique Fédérale de Lausanne (EPFL), Lausanne, Switzerland}

  \icmlcorrespondingauthor{Simon Guilloud}{simon.guilloud@epfl.ch}
  \icmlcorrespondingauthor{Sankalp Gambhir}{sankalp.gambhir@epfl.ch}
  \icmlcorrespondingauthor{Samuel Chassot}{samuel.chassot@epfl.ch}

  \icmlkeywords{keywords}

  \vskip 0.3in
]

\printAffiliationsAndNotice{All authors contributed equally.} 

\begin{abstract}
  We present a case study in \emph{reformalization}, a variant of autoformalization in which the input proof is not natural language but a formal development in a different proof assistant. Concretely, we report three reformalizations of the Jordan Curve Theorem: from Mizar to Lean, from HOL Light to Lean, and from HOL Light to Agda. We analyse the results and identify pipeline design choices that matter for practical reformalization tasks. The final formalizations are available at \url{https://github.com/epfl-lara/jordan-curve-theorem}.
\end{abstract}

\section{Introduction}
Much attention has been given to the problem of autoformalization: given a mathematical text in natural language (such as a paper or textbook), produce a corresponding formalization in a proof assistant. Much less attention has been given to the simpler but still highly useful question of formalizing a mathematical theory not from an informal proof, but from a development in a different proof assistant. We call this task \textit{reformalization}.

Automated proof transfer, which translates proofs across proof systems, has long been a research topic in the ITP community \cite{kohlhaseExperiencesExportingMajor2021,anonymousQEDManifesto1994}. Before recent progress by LLMs in formal mathematics, the main viable approach was low-level logical translation of statements and proofs. Many approaches were developed over the years, but they face two fundamental obstacles. First, a mismatch between logical foundations (for example, dependent type theory, higher-order logic, or set theory) is difficult to overcome. Even when the logical foundations are provably equivalent, existing approaches are typically restricted to transfers between systems with very similar foundations, transfers from a simpler system to a more complex system (for example, \cite{blanquiTranslatingHOLLightProofs2024,kellerImportingHOLLight2010}), or translations with large complexity and proof-size overhead, as in \cite{guilloudMechanizedHOLReasoning2024}.

The second obstacle is that, even when technically feasible, the resulting translation is unnatural. In particular, the translation of a concept often does not match the way that concept is defined in the target assistant. This is the problem of \textit{library alignment}. For example, in a transfer of HOL Light proofs into Lean, the natural numbers $\mathbb N$ as defined in HOL Light and translated into Lean yield a definition of $\mathbb N$ that is not only different from Lean's inductive type $\mathbb N$, but possibly not even provably equivalent to it in Lean. The usual solution is manual alignment of concepts, which restricts the use of each translated concept to a particular API\footnote{For example, in set theory, $0$ is often defined as the empty set $\emptyset$. But the theorem $0 = \emptyset$ certainly cannot hold (or even be stated) about the number $0$ as defined in, e.g., Lean. Such ``accidental'' theorems cannot be used in a translation that maps the set $\mathbb N$ to the Lean type $\mathbb N$.}.

In mathematical research, however, and with the exception of a few peculiar domains, it is widely understood that the particular choice of logical foundations matters little: in principle, a mathematical proof can be carried out ``naturally'' in any proof system with sufficient background. In that regard, the two problems above are artefacts of translations that must be deterministic, complete, and correct by construction, but that cannot preserve the spirit of the original proof. LLM-based proof transfer is the opposite: it is not deterministic, complete, or correct by construction, but it can preserve the proof idea and thereby largely avoid the two obstacles above. This is the rationale behind \textit{reformalization}.

In the present work, we show through a case study that suitable agentic pipelines combined with large language models can transfer large mathematical developments across entirely different proof assistants. Concretely, we consider the Jordan Curve Theorem (JCT) as our main study. The JCT states that any non-self-crossing closed curve separates the plane $\mathbb R^2$ into two disjoint regions -- a bounded interior and an unbounded exterior -- and was proven by Camille Jordan in 1887. It is a famous example of an apparently obvious theorem whose proof is long and technical, and it has important applications.

The first formalization of the JCT was finished in 2005 by Hales \cite{hales2007jordan} in the HOL Light proof assistant \cite{harrisonHOLLightOverview2009}, in a proof of roughly 60,000 lines. In the same year, an international team finished another formalization of the theorem in the Mizar proof assistant \cite{kornilowicz2007proof}. The JCT proof effort in Mizar spread over 14 years and, because of the attraction of the problem and the tendency of those who undertake it to disappear, has been described as a black hole into which formalizing mathematicians fall, never to be seen again. Although Hales was faster, he still describes his formalization effort as having taken ``many months of hard work''. With an existing formal proof as baseline, manually porting the proof to a different proof assistant would take less time, perhaps ``a few'' months of hard work. In contrast, the reformalizations in the present case study produced JCT formalizations in about a week, with no more than two days of active human involvement and using commercially available LLMs. This demonstrates that practical proof transfer is now feasible at scale.

In the rest of the paper, we present three reformalizations of the JCT: from Mizar to Lean, from HOL Light to Lean, and from HOL Light to Agda. 
Lean \cite{demouraLeanTheoremProver2015} was chosen for its well-developed ecosystem of tooling and the Mathlib \cite{mathlib2020} library, and Agda \cite{Norell2007} was chosen as an example of a proof assistant with similar foundations, but less developed tooling in comparison, to test the impact of the prover's ecosystem.
For each, we present the pipeline and report the design choices that affected the agent's performance. We then compare the three reformalizations to evaluate how much the source and target proof assistants, and in particular their foundations and tooling, matter for practical reformalization.

The formalizations are available together under an Apache 2.0 licence at \url{https://github.com/epfl-lara/jordan-curve-theorem}.

\section{Reformalization Task and Evaluation Setup}

We use the term \emph{reformalization} for the following task: given an existing formal development establishing a statement $T$, produce a new, independently checkable formal development in a different target proof assistant that establishes the corresponding claim $T'$. In contrast to correct-by-construction proof translation, our goal is not to preserve source syntax, proof terms, or definitional equality, but to recover in the target assistant a proof development that is mathematically faithful to the source and idiomatic with respect to the target.
Hence the source formalization is not treated as an object to be compiled, but as an exact specification of the theorem and a highly detailed, guaranteed-correct proof. The LLM-based agent uses this specification together with the target assistant's libraries, search tools, and a verification loop to construct a native development for the target proof system.

\paragraph{Scope of the task.}
Our case study focuses on the Jordan Curve Theorem and its supporting infrastructure. This is a useful stress test for reformalization for three reasons. 
First, the statement of the theorem only requires real numbers and continuous functions, and hence can be stated directly in all the assistants we consider. Second, the proof developments are large and technically heterogeneous, mixing topology, geometry, graph arguments, and analytic lemmas. Third, mature source formalizations exist in more than one assistant, allowing us to compare reformalization across different source languages and libraries.

For each case study, the setup consists of: 
\begin{itemize}
  \item a source formalization of the Jordan Curve Theorem,
  \item a target assistant together with its standard library and existing ecosystem tools, and
  \item an agentic pipeline combining an LLM, tool use, and human-written prompts and directives.
\end{itemize}

\paragraph{Human role.}
No proof code was written by hand, and no mathematical insight or proof guidance was given to the agent. Human input was used to configure the workflow: prompts were revised, additional verification steps were inserted, and observed failure patterns were used to improve the process. This is precisely the kind of engineering experience we want to surface in the paper.

\paragraph{Evaluation goals.}
Our evaluation is designed to answer three questions. First, can current LLM-based pipelines transfer a large, nontrivial formalization across proof assistants while keeping human effort low? Second, which factors most affect practical success: the quality of the tooling of source and target assistants, the maturity of their libraries, the nature and mismatch of their logical foundations, the model used, or the design of the agentic pipeline? Third, what are good agentic pipelines and guardrail designs for reformalization?

\paragraph{Measured quantities and reported observations.}
A reformalization counts as a verified success if the resulting target development is fully verified in the target assistant (without \texttt{sorry} placeholders or additional axioms) and establishes the final Jordan Curve Theorem statement.
For each reformalization, we first give a qualitative experience report centred on pipeline design. In
particular, we emphasize the practically useful lessons we learned about how to structure the workflow, where
agents fail, and which design choices and guardrails make reformalization succeed.

We then complement this experience report with quantitative descriptors: estimated human involvement time, size of
the resulting development (number of theorems, number of lines, number of tactics), compilation time, the
proportion of the original proof that was already present in the target assistant's library versus the part that
had to be actively formalized, the number of active days needed to complete the project, and approximate LLM cost.
These metrics are admittedly imperfect -- proof size is a mediocre proxy for complexity, and time and cost are
largely vendor-dependent -- so they should not be understood as a definitive benchmark, but rather as supporting
context for the qualitative report.

\section{JCT: Mizar to Lean}

The reformalization of JCT from the Mizar Mathematical Library (MML) was
conducted in three distinct steps: (1) extraction of dependency information,
(2) construction of a proof sketch and library alignment, and (3) proof filling. We used both the original Mizar tool \cite{naumowiczBriefOverviewMizar2009} and the \texttt{mizar-rs} proof checker reimplementation \cite{carneiro:LIPIcs.ITP.2023.10}.

\paragraph*{Step 1: Extracting Metadata.}
The Mizar proof checker is not a single tool, but rather a collection of tools
that execute separately at different stages and exchange data through standardized intermediate
text formats. Unlike a closed environment where proof data is stored entirely in memory,
the text formats enable independent processing and analysis of the proof data.

We used the intermediate files generated by a successful run of the proof
checker to extract dependency graphs between source files as well as theorems in
the development leading to the final Jordan Curve Theorem statement.


\paragraph*{Step 2: Proof Sketch and Library Alignment.}
The dependency graph and MML sources were provided to an agent with instructions
to produce one Lean file per Mizar source file, translating Mizar theorems into
Lean theorem skeletons, and inserting comments with the original Mizar source
element, location, and the theorems and lemmas used in the current theorem's proof.

In a second pass, the agent was instructed to augment the metadata with the closest 
matching theorems existing in Mathlib for each component imported from Mizar.

The resulting statement of the Jordan Curve Theorem was then manually verified
to be correct and to only rely on definitions from Mathlib.


\paragraph*{Step 3: Proof Filling.}
The agent was finally instructed to fill in the proof of each theorem, utilizing
the metadata and Mizar sources. The Mizar source provided the proof sketch, while
the metadata provided a closed set of lemmas that could be used in the proof. The 
agent was instructed to delegate to Mathlib whenever the closest match found in 
the previous step was sufficient to close the goal.

Despite the metadata and source links, the agent sometimes
diverged by attempting to write a proof sketch from scratch, particularly
following a context-window summarization. In this case, the agent was manually
steered to refer to the Mizar sources. This occurred approximately 10 times over
the course of the formalization, but otherwise no mathematical guidance was
provided.

Progress was tracked through a status file auto-generated using a script
filtering the Lean build output, and a markdown file containing a human-readable
formalization plan, which the agent was instructed to update after each
completed theorem.

\subsection{Results}


The final Lean development contains 413 lemmas and theorems, spanning
approximately 474,000 characters (about 11,000 lines). In comparison, the source
Mizar development (restricted to \texttt{JORDAN*}, \texttt{BROUWER*}, and
\texttt{BORSUK*}) spans approximately 775,000 characters (about 23,500 lines).
The difference in size can be largely attributed to lemmas being discharged by
close matches in Mathlib. The project compiles in 111s (wall-clock time for \texttt{lake
build}), with Mathlib cached.

The full formalization took about 10 days to complete, using Claude Opus 4.6 and
GPT-5.4 via GitHub Copilot. We estimate the human involvement, which consisted
of setting up the project, monitoring progress, and steering the agent, at
about 20 hours.

\paragraph*{Divergence from Mizar.}
Because the pipeline prioritized existing Mathlib lemmas whenever they were
sufficient, the resulting Lean development did not always reproduce Mizar's
internal proof machinery. The overall structure of the Jordan Curve Theorem
argument remained close to the source, but several intermediate constructions
were replaced by more idiomatic Mathlib-based reasoning. In particular,
substantial parts of the Mizar Go-board proof engineering were no longer needed,
as equivalent topological results were already available in Mathlib.

\section{JCT: HOL Light to Lean}
\label{sec:hol-to-lean}
In contrast with the Mizar-to-Lean formalization, the formalization from HOL Light to Lean was linear. The HOL Light proof of the Jordan Curve Theorem is split into 30 sections (labelled A to DD). The agent was tasked with working section by section and theorem by theorem, without computing a dependency graph or prereading future sections. The agent's workflow for each section was set up as follows:
\begin{enumerate}
    \item Populate a catalogue of HOL Light theorems and definitions with a dedicated script.
    \item Build a Lean skeleton of theorems and definitions, using \texttt{sorry} placeholders for proofs.
    \item Fill all sorries one by one, in order, building between each.
    \item Once the file is \texttt{sorry}-free and compiles cleanly, update the catalogue with the corresponding Lean theorem and update the status document.
    \item Read the instruction file again and move on to the next section.
\end{enumerate}

\paragraph{Populate the Catalogue.}
After reading a status file to confirm the project's state and current section, the agent executes a dedicated Python script that extracts all theorems and definitions from a given section of the HOL Light proof and adds them to a CSV catalogue. Each entry contains the theorem's name, its section, its kind (definition or theorem), its formal statement, and whether it is used beyond the section itself or only locally. Columns for the corresponding Lean statements are filled afterwards.

\paragraph{Write the Skeleton.}
The agent must then write a skeleton of the section in a Lean file, with \texttt{sorry} placeholders for each theorem. Every theorem or definition has to be classified in one of the following three categories:
    \begin{itemize}
        \item \emph{Exists in Mathlib.} The theorem is covered by an existing theorem from Mathlib. The agent must add the theorem name in Mathlib to a dedicated column in the CSV.
        \item \emph{Not applicable to Lean.} Some theorems genuinely do not apply in Lean, typically because they are immediate by type-checking. This must also be recorded in the CSV.
        \item \emph{Must be implemented.} The theorem or definition must be implemented. The theorem name in Lean and its file and line numbers must be updated in the CSV.
    \end{itemize}
In early exploration, the agent was given less strict guidelines. It sometimes decided that a theorem could be skipped, for example because the proof was simple and could be inlined, or because it had found a way to carry the proof without using the theorem. However, on multiple occasions this led the agent to become stuck on a theorem in a later section because a theorem was missing or a key lemma had a weaker statement than its HOL Light counterpart.

This motivated strict maintenance of the CSV catalogue, and the agent was given a strict directive to implement \emph{every} theorem that was not in Mathlib or made trivial by Lean type checking.

\paragraph{Fill proofs.}
The agent is then tasked with filling proofs one by one, in order. It always has to first read the HOL Light proof to understand the strategy, and then implement it using the already formalized theorems and definitions. It builds and verifies compilation before moving to the next \texttt{sorry}, and was explicitly forbidden to work on multiple proofs at once.

Longer proofs in HOL Light often triggered the agent to search for shortcuts: its visible behaviour suggested an aversion to long, complicated tasks. But the resulting search for a shorter proof typically took much longer than the time required to transport the proof as-is.
Hence, the agent was given the guideline to always strictly follow the HOL Light proof and not to search for shortcuts. It was also told that the HOL Light proofs were already optimized and that the important thing was to keep making progress. Limiting the agent's ``creative freedom'' and imposing formal, mechanical bookkeeping made the process much smoother and faster.

\paragraph{Verify and advance.} Finally, once the file compiles, contains no remaining \texttt{sorry} placeholders, and only uses standard axioms, the agent should update the CSV and the status file, then move on to the next section.

After completion, we manually verified that the proven statement was indeed a correct formalization of the JCT and then had the agent proof-golf, refactor, and improve the quality of the development. We used a second agent, a Lean reviewer given Mathlib's guidelines for PR quality, and had the two agents engage in a few rounds of proof improvement.

\subsection{Results}

The final Lean formalization comprises close to 1,400,000 characters (29,000 lines), against 1,709,000 characters (59,000 lines) in the HOL Light proof. It contains more than 910 theorems and lemmas and 150 definitions. Of these, 116 have no equivalent in HOL Light: they are helper lemmas or formalize statements from the HOL Light library that did not exist in Mathlib. On the other hand, 258 statements from the HOL Light proof were already in Mathlib, and 213 are classified as ``not applicable to Lean''. We observe that, once corrected for the difference in theorems on each side, the number of characters in each proof is quite similar, meaning that both proofs are similarly verbose in their respective languages. The project takes 206s to compile (wall-clock time for \texttt{lake build}).

The full formalization took a week to complete, using Claude Opus 4.6 via GitHub Copilot. We estimate the total active human involvement at about 10 hours; this consisted of setting up the project and the agentic pipeline, as well as monitoring the agent's progress. The pipeline first evolved according to observations of the walls the agent encountered. However, from around the middle of the formalization (section N), no further modifications were needed, and the agent completed the second half entirely autonomously.

\section{JCT: HOL Light to Agda}

To evaluate the importance of the target proof assistant in such agent-based proof translation, we apply the same approach to port the Jordan Curve Theorem proof to Agda~\cite{Norell2007} from HOL Light. 

We used the same protocol as for the Lean target, explained in Section~\ref{sec:hol-to-lean}. We adapted the instructions to target Agda instead of Lean and to use the appropriate terminology. We used Claude Opus 4.7 via Claude Code CLI for this instruction adaptation and for the formalization itself.

At the time of writing, the Agda formalization is not complete. Section 0 (i.e., the foundations), as well as Sections A to D and around half of section E are ported and compile. This amounts to around 27,000 lines of Agda code. As the main theorem is not ported yet, we cannot confidently assess the correctness of the proof porting. We therefore report on the experience in setting up such experiments with Agda, and lessons we learned while doing so.

\paragraph{Empty types.} At the beginning of Section 0, the agent asked the human operator multiple questions regarding empty types. In HOL Light, types are strictly inhabited, while in Agda types can be empty. The main consequence, as the agent pointed out, is that some total functions in the HOL Light formalization are not total as-is in Agda. The agent suggested different options, such as assuming that types are nonempty (which is unsound) or threading evidence of an inhabitant through the definitions that need it. We told the agent to use the latter.

\paragraph{Axiomatization.} Also during the formalization of Section 0, the agent asked the human operator whether it should axiomatize some theories or build them from the foundations. We told the agent to build them from scratch for added robustness, at the expense of time.

\paragraph{Memory pressure.} During the port of Section 0, Agda type checking exceeded the memory available to the Docker container where the agent was running (32GB). The agent did not realize this and interpreted the absence of a reply from the executable as successful type checking, while Agda had in fact been killed by the operating system for running out of memory (OOM). The human operator had to point this out before the agent understood the failure mode and took action. Fixing the proofs to type check within the available memory took days. The Agda type checker was running for 5 to 10 minutes before being killed by the operating system, which prevented fast agent turnaround. The agent finally succeeded by splitting the proofs into chunks small enough to type check within the available memory. During this process, we also updated the instructions to force the agent to use Agda's cached interfaces to speed up type checking of subsequent chunks. Using the cache was particularly critical given the runtime of the type checker on the most demanding proof chunks.

\subsection{Takeaways}
The Agda experiment suggests that the target assistant's operational characteristics matter as much as its foundations. The main technical mismatch with HOL Light was not only the difference between higher-order logic and dependent type theory, but also the need to make HOL Light's inhabited-type discipline explicit. The main tooling issue was turnaround time: once type checking required minutes and could fail by OOM without a clear diagnostic to the agent, progress slowed sharply. For Agda, robust reformalization pipelines should therefore make non-emptiness assumptions explicit, avoid unnecessary axiomatization when possible, split developments aggressively, and enforce cached type-checking workflows from the beginning.

To make the most out of the LLM agents, the verification/compilation process should be as fast as possible. This takeaway is not exclusive to LLM agents as human users also benefit from fast verification. However, LLM agents tend to follow a trial-and-error approach with fast turnaround to writing proofs.

Caching verified results is crucial, especially with tools with slower verification. Indeed, almost instant verification is not always possible, but the agent should not wait for already verified results, for the same turnaround reason. The human operator should also ensure that instructions are clear on that matter.

The tool should also provide as much feedback as possible. Again, this matter is not exclusive to LLM agents, but they tend to lack the intuition of experienced human users.

Finally, instructions should be clear about the fact that the verification might fail for external reasons such as getting OOM killed by the OS. The agent must therefore take that possibility into account to avoid missing critical feedback.

\section{Comparison to automated proof transfer}

The problem of translating proofs across different proof assistants has been
studied in the ITP community. The standard approach is to implement a general
translation that transfers low-level proofs from one proof assistant and checks
them with another.
The approach typically requires high engineering
effort and matching logical foundations, and the
resulting translated proofs lack good library alignment and are often not
idiomatic with respect to the target system. 

In particular, we compared our HOL Light to Lean reformalization against prior
work on automated proof transfer between HOL Light and Isabelle/HOL
\cite{kaliszyk2013scalable}. The process patches HOL Light to generate proof
trees in a serialized format. The outcome of the pipeline is an Isabelle theory
file which loads the standard library shipping with the translation, which
matches some of the core concepts from HOL Light to Isabelle/HOL. Then, the
script delegates to an OCaml program that reads the exported proof trees and
checks them within the Isabelle kernel. The result is a verified translation of
the Jordan Curve Theorem, but one which cannot easily fit into Isabelle's large
existing mathematical library, does not reuse any existing non-trivial
components, and does not produce a human-readable Isabelle theorem.

While this is logically sufficient to verify the theorem, it significantly
limits the usefulness of such a translation for real-world use. In contrast,
LLM-based reformalization is not deterministic, complete, or correct by
construction, but it can preserve the original arguments rather than
low-level proof terms, leverage existing libraries, and tune the resulting
definitions and proofs to be better integrated with the target proof assistant's
ecosystem. Correctness is preserved by utilizing the target assistant's proof
checker as the final arbiter.

\section{Conclusion}

Our case study shows that LLM-based reformalization can transfer large formal developments across proof assistants at a scale that would be impractical for manual porting and difficult for correct-by-construction proof translation. The development of mathematical libraries has always been a major bottleneck in the development of proof assistants, and we believe that this will be radically improved by reformalization.

Nonetheless, LLM-based reformalization is not a silver bullet, and it does not readily address heavy mismatches in logical foundations. In all our case studies, proof assistants were already well developed, and, additionally, the Jordan Curve Theorem is largely independent of their particular foundations. We would expect naturally translating theorems in, for example, set theory, which is highly dependent on particular encodings, to be more difficult for agentic systems. Hence, we expect that in the future reformalization tools will integrate elements of systematic, logic-based approaches. 

Our case study also highlights important takeaways for reformalization pipelines. We highlight the importance of a well-defined workflow in the agent instructions, including source-aware decomposition for a clear idea of progress, strict bookkeeping for easy backtracking, alignment with the existing library, and guardrails that prevent the agent from abandoning the source proof and designing its own arguments. In contrast with autoformalization, where natural language proofs often contain implicit arguments or gaps, formal proofs are necessarily fully detailed and correct. This makes the process fully mechanical, so limiting the agent's freedom was key in improving performance. The comparison across Mizar, HOL Light, Lean, and Agda suggests that libraries and tooling strongly affect success: Lacking libraries need to be completed first by the agent, and slow compilation processes can drastically slow progress. On the other hand, foundational mismatch can often be handled when the workflow makes the mismatch explicit and keeps the target proof checker in the loop.


\section*{Impact Statement}
This paper presents work whose goal is to advance the state of the art in leveraging artificial intelligence for mathematical research. There are many potential societal consequences of our work, none of which we feel must be specifically highlighted here.

\bibliography{sguilloud, biblio}
\bibliographystyle{icml2026}

\end{document}